\documentclass[letterpaper]{article}
\usepackage{aaai20}
\usepackage{times}
\usepackage{helvet}
\usepackage{courier}
\usepackage[hyphens]{url}
\usepackage{graphicx}
\usepackage{graphicx}
\usepackage{bbold}
\usepackage{amsmath}
\usepackage{amssymb}
\DeclareMathOperator*{\argmin}{\arg\!\min}
\newcommand{\citet}[1]{\citeauthor{#1}\shortcite{#1}}
\newcommand{\citep}{\cite}

\title{EMAP: Explanation by Minimal Adversarial Perturbation}
\author{Matt Chapman-Rounds\\
  Department of Informatics\\
  Edinburgh University\\
  Edinburgh, EH8 9AB \\
  \texttt{m.rounds@ed.ac.uk} \\
  \And
  Marc-Andre Schulz \\
  Department of Psychiatry, Psychotherapy and Psychosomatics  \\
  RWTH Aachen University \\
  Aachen, Germany \\
  \texttt{marc.schulz@rwth-aachen.de} \\
  \AND
  Erik Pazos \\
  QuantumBlack \\
  100 Museum Street, London, WC1A 1PB \\
  London, SW1Y 5AU \\
  \texttt{erik.pazos@quantumblack.com} \\
  \And
  Konstantinos Georgatzis \\
  QuantumBlack \\
  100 Museum Street, London, WC1A 1PB \\
  London, SW1Y 5AU \\
  \texttt{konstantinos.georgatzis@quantumblack.com}
  }

\begin{document}
\maketitle
\begin{abstract}
Modern instance-based model-agnostic explanation methods (LIME, SHAP, L2X) are of great use in data-heavy industries for model diagnostics, and for end-user explanations. These methods generally return either a \textit{weighting} or \textit{subset} of input features as an explanation of the classification of an instance. An alternative literature argues instead that \textit{counterfactual} instances provide a more useable characterisation of a black box classifier's decisions. We present EMAP, a neural network based approach which returns as Explanation the Minimal Adversarial Perturbation to an instance required to cause the underlying black box model to missclassify. We show that this approach combines the two paradigms, recovering the output of feature-weighting methods in continuous feature spaces, whilst also indicating the direction in which the nearest counterfactuals can be found. Our method also provides an implicit confidence estimate in its own explanations, adding a clarity to model diagnostics other methods lack. Additionally, EMAP improves upon the speed of sampling-based methods such as LIME by an order of magnitude, allowing for model explanations in time-critical applications, or at the dataset level, where sampling-based methods are infeasible. We extend our approach to categorical features using a partitioned Gumbel layer, and demonstrate its efficacy on several standard datasets. 
\end{abstract}

\section{Introduction}

Recent interest in explaining the output of complex machine learning models has been characterized by a wide range of approaches \citep{lipton2016mythos,MontavonSM17}. Many of these approaches are model specific; for example attempts to explain neural networks that rely on interpreting the flow of gradient information through the model \citep{shrikumar2017learning,olah2017feature,karpathy2015visualizing}, or decision trees, which might be considered directly interpretable \citep{molnar2019}. 

Model agnostic approaches, however, are attempts to formulate a general framework for per-instance explanation of a model's outputs regardless of the type of model being used. This can be beneficial both in circumstances where choice of model may change over time, or where the original model is costly or impossible to access. 

One group of model-agnostic explainers focuses on providing an explanation of a model's output as either a subset of input features \citep{anchors:aaai18,chen2018learning}, or a weighting of input features \citep{ribeiro2016should,lundberg2017unified} of the instance to be explained. Another group of models \citep{white2019measurable,wachter2017counterfactual} instead proposes that counterfactual instances, or groups of instances, are a useful proxy to 'explanation', where the claim is that local explanations are expected to contain both the outcome of a prediction, and how that prediction would change if the input changed. Many of these approaches use sampling procedures to either estimate local decision boundaries (and their corresponding parameters), or to find proximate counterfactual instances, and are thus computationally expensive. The computational cost of sampling the local decision boundaries for each new explanation makes these methods slow to scale, and of limited use in real-world applications.

We propose EMAP, Explanation by Minimal Adversarial Perturbation, a model that returns the direction that an instance would have to be perturbed the least in order for the classification of the underlying model to change. EMAP's contribution is threefold: 

\begin{itemize}
    \item EMAP combines elements of both the feature weighting and counterfactual paradigms of model explanation, and is fully model-agnostic.
    \item EMAP is faster than alternative methods by 5 orders of magnitude, once constant overheads are taken into account, allowing for model explanations in time-critical applications where sampling-based methods are infeasible.
    \item EMAP naturally indicates regions of low classifier confidence, or potential user interest, as a consequence of its design.
\end{itemize}{}

The paper is structured as follows. In Related Work we provide an summary of recent alternative approaches to instance-wise model-agnostic explanation. The Model section includes a justification of our approach, and separately describes how we handle continuous and categorical input variables. Experimental Setup details the datasets and model choices made. The Results section analyses the model's performance on two synthetic datasets with continuous features, a more complex continuous dataset, and a standard dataset with both continuous and categorical features. We summarise our findings in the Conclusion.

\section{Related Work}

One of the most widely-used feature weighting approaches to per-instance explanation of a black box model's outputs is LIME \citep{ribeiro2016should}, which learns a local surrogate approximation to the original model's output, centered on the instance to be explained. It does this by generating a new dataset of permuted samples and corresponding predictions of the black box model, and then trains an interpretable linear model on this new dataset, where each point is weighted by its proximity to the point of interest. The weights of the linear model are then considered to be the explanations of the black box model's output at that point. LIME can also be considered to be slow; its reliance on sampling afresh for every data point reduces the speed at which explanations can be collected for large numbers of instances of interest.


Separate work has shown that those explanation methods that return a weighting of input features, including LIME, can all be considered as additive feature attribution methods, with an explanation model that is a linear function of binary variables \citep{lundberg2017unified}. This unified framework is called SHAP, and accompanying methods exist to estimate feature importance values for instance predictions on particular models \citep{lundberg2018consistent}.

One attempt to produce \textit{fast} instance-based explanations is L2X \citep{chen2018learning}, where the authors train a neural network to output a binary mask over instance features, and a second network to return the original black box model output from the masked input. By training on a cross entropy objective, they argue that they are effectively maximising the mutual information between some subset of input features and the true model output. The subset of features chosen once the explainer is trained should be the maximally informative subset, and thus a good explanation of the black box model output. This approach shares some similarity with ours, insofar as the second network can be thought of as learning a differentiable surrogate to the true model, although the authors do not consider their model in these terms. A crucial drawback of L2X is that it does not provide weighting of feature importances, nor does it provide the direction in which a given feature would impact classification.

An example of the fact that adversarial examples can be good explanations of underlying models is the work of \citet{wachter2017counterfactual}. Here the approach, assuming a trained model $f_w(x)$, is to minimise
\begin{equation*}
    L(x,x',y',\lambda) = \lambda (f_w(x') - y')^2 + d(x,x')\ ,
\end{equation*}
where the first term is the quadratic distance between the output of the model under some counterfactual input $x'$ and a new target $y'$, and the second term is a measure of the distance between the true input to be explained, $x$, and its possible counterfactual instance $x'$. This approach is similar in spirit to ours, but differs in several important ways. 

Firstly the method returns a set of counterfactual instances, rather than a counterfactual direction. Secondly, the procedure to generate one counterfactual example for one point requires iterating between minimising the above objective and increasing $\lambda$, and the authors recommend initialising a sample of potential counterfactuals and repeating the process on all of them, to avoid getting stuck in local minima. This means the process is slow. Thirdly, optimising the above objective assumes that $f_w(x)$ is tractable (for example, a gradient based optimiser would need the gradient of $f_w(x)$ with respect to $x$). This limits the approach to only those models where this is the case, whereas by training a differentiable approximation to the black box model, we circumvent this issue.

Another similar approach can be found in CLEAR, \citep{white2019measurable}, which includes an interesting model of fidelity, although again the process of extracting an explanation requires sampling, and iterative solving.

In short, LIME, SHAP, and other sampling-based models require thousands of model-evaluations for each instance that needs to be explained. L2X needs only one forward pass of a neural network per explanation, but does not provide a weighting of feature importances, nor directionality of explanations. With EMAP, we provide a method that retains the benefits of LIME and SHAP, while providing computational efficiency on par with L2X.

In the domain of explaining the outputs of neural networks, particularly for image classification, there are several examples of papers which use adversarial or perturbatory approaches \citep{dabkowski2017real,dhurandhar2018explanations,fong2017interpretable}. These approaches often rely on dividing images into regions, which places a strong modelling prior on correlations between input features (in this case, pixels). As our approach is fundamentally more general, we are not able to make similar assumptions, and likely would have substantially different use-cases. Two of these papers \citep{dabkowski2017real,dhurandhar2018explanations} also assume differentiability, whilst the third treats 'perturbations' as one of three regional noise masks; instead of learning feature-specific meaningful perturbations as in our approach.

\section{Model}

\subsection{Overview}

Our general approach to the problem of explaining an instance's classification by a model is to find the minimal \textit{adversarial} perturbation of that instance. This can be thought of as an answer to the question 'what is the smallest change we can make to this instance to change its classification?'. We argue that this is a useful measure for two reasons.

First, it is locally meaningful. An instance's classification depends on its location relative to the classifier's decision boundary or boundaries. The minimal adversarial perturbation will 'point' directly to the nearest decision boundary. Features that contribute substantially to this minimal perturbation must also be features that have contributed substantially to the instance's classification. If we imagine perturbing the features of an instance equally, those with relatively large contributions to the original classification will be just those that have a relatively large contribution to subsequent misclassification.

Secondly, it is useful for an end-user. The outputs of a model often require explanation due to a desire for improvement, or, more specifically, instances that require further justification are often instances which have been wrongly classified, or are suspected to have been wrongly classified. Indicating what should be changed to allow an instance to be alternatively classified satisfies this requirement directly, and in a manner which is arguably more interpretable than providing the weights of a local linear model.

\subsection{Continuous input features}
Let us assume we have access to a set of outputs $\{f(\pmb{x}^{(n)})\}^{N}_{n=1}$ of some model $f: \mathbb{R}^d \to \mathbb{R}$, where for binary classification, $f(\pmb{x}^{(n)})$ will be the probability that input instance $\pmb{x}^{(n)}$ belongs to the target class \footnote{For the sake of clarity, we will initially assume a binary classification. Multi-class classification is dealt with below, and regression is discussed in the conclusion}, or a corresponding indicator function ($\mathbb{1}[f(\pmb{x}^{(n)})] = 1, f(\pmb{x}^{(n)}) \geq 0.5$). For each $\pmb{x}^{(n)}$ we wish to explain, our goal is to find the smallest adversarial perturbation; i.e. the smallest perturbation $\pmb{p}^{(n)}$ such that $\mathbb{1}[f(\pmb{x}^{(n)})] = 1 - \mathbb{1}[f(\pmb{x}^{(n)} + \pmb{p}^{(n)})] $. Here, $\pmb{p}^{(n)} \in \mathbb{R}^d$, and if minimal, can be thought of as the shortest distance from $\pmb{x}^{(n)}$ to the decision boundary of $f$.
\par The space of possible perturbations $P$ is prohibitively large for an exhaustive search per instance to be explained, and so we will assume a restricted class of models $G: X \to P$, mapping data space to perturbations. Our approach in this paper is to represent such a mapping as $g(\pmb x; \theta_g): \mathbb{R}^d \to \mathbb{R}^d $, $g \in G$, a differentiable function described by a neural network with parameters $\theta_g$. Ideally, we would then like to compute the optimal adversarial parameter settings $\hat{\theta}_g$ by standard gradient-based methods, using:

\begin{equation}
\begin{aligned}
    \hat{\theta}_g = \argmin_{\theta_g} \Big\{ - &\sum^{N}_{n=1}\Big( 1-\mathbb{1}[f(\pmb x^{(n)})]\Big)\log f\Big(\pmb x^{(n)} \\ 
    &+ g(\pmb x^{(n)};\theta_g)\Big) + \lambda|g(\pmb x^{(n)};\theta_g)|_2 \Big\} \ ,
\end{aligned}
\end{equation}
where $\lambda$ is a hyperparameter restricting the size of generated perturbations, and $1-\mathbb{1}[f(\pmb x^{(n)})]$ are the adversarial labels.
\par However, in a model-agnostic setting, we cannot assume $f$ to be differentiable\footnote{Or at least, we cannot assume we have access to the gradients of $f$.}, or even that we have access to $f$ itself to compute $f(\pmb x^{(n)} + g(\pmb x^{(n)};\theta_g))$. We therefore further define a surrogate $s(\pmb x; \theta_s): \mathbb{R}^d \to \mathbb{R}$, also a neural network, which is trained to be a differentiable approximation to $f$ by cross entropy loss:

\begin{equation}
    \hat{\theta}_s = \argmin_{\theta_s}-
    \sum^{N}_{n=1} \mathbb{1}[f(\pmb x^{(n)})]\log s(\pmb x^{(n)}; \theta_s)\ .
\end{equation}
Substituting $s(x; \hat{\theta}_s)$ for $f$ in (1) finally gives us a tractable objective:

\begin{equation}
\begin{aligned}
    \hat{\theta}_g = \argmin_{\theta_g}\Big\{ - &\sum^{N}_{n=1}\Big( 1-\mathbb{1}[f(\pmb x^{(n)})]\Big)\log s\Big(\pmb x^{(n)}\\ 
    &+ g(\pmb x^{(n)};\theta_g); \hat{\theta}_s\Big) + \lambda|g(\pmb x^{(n)};\theta_g)|_2 \Big\} \ .
\end{aligned}
\end{equation}
Note that $\mathbb{1}[f(\pmb x^{(n)})]$ remains unchanged, as it does not depend on $\theta_g$, and we have assumed we know $f(\pmb x^{(n)})$ for all $\pmb x^{(n)}$ in our data. 
\par In practice, training is carried out in two stages; firstly we train $s(\pmb x;\theta_s)$ on the original inputs and original labels to approximate the black box model $f$. Secondly, we freeze the weights of $s$ and train $g(\pmb x; \theta_g)$ on the original inputs and flipped labels; the perturbations $\pmb p^{(n)}$ output by $g(\pmb x^{(n)}; \theta_g)$ are added to the original inputs and passed through the surrogate $s$. As $s$ is a differentiable model, back-propagation provides the gradients of the loss with respect to the perturbations, and hence with respect to $\theta_g$. We can therefore train $g$ directly using the original dataset.

\subsection{Discrete input features}
For many applications, however, some or all of the input features of $f$ will be discrete, rather than continuous. For some categorical feature $x_i$, which takes values $\{1,...,K\}$ outputting a continuous value $p_i$ from our perturbation generator $g$ is unhelpful. We first consider the case in which all input features are categorical.

We take the general approach that perturbing a categorical feature means sampling from a corresponding categorical distribution and assigning the feature the sampled value. For each categorical $x_i$, our mapping $g$ from data space to perturbation space contains the corresponding sub-mapping $g_i(x^{(n)}_i, \theta_g): \mathbb{R}^K \to \mathbb{R}^K$, assuming a 1-hot encoding, where each of the $K$ real valued outputs is treated as the log class probability $\log \pi_k$ of the $k^{th}$ value of the categorical feature.
\par To train to find adversarial samples, we can use the softmax function as a continuous differentiable approximation to $argmax$, which allows us to use the Gumbel-Softmax trick to generate $K$-dimensional sample vectors $y$ where the $k^{th}$ element is given by:

\begin{equation}
    y_k = \frac{\exp((\log \pi_k + g_k)/\tau)}{\sum_{j=1}^K \exp((\log \pi_j + g_j)/\tau)}\ ,
\end{equation}

\begin{figure*}[t]
    \centering
    \includegraphics[width=\textwidth]{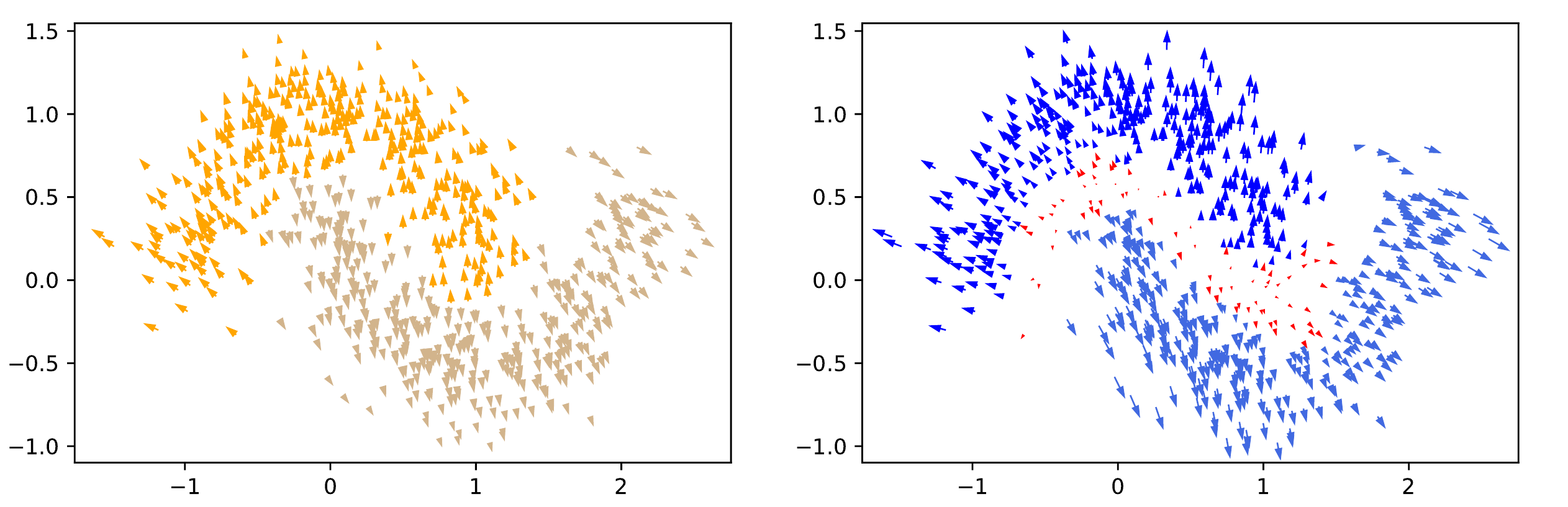}
    \caption{Performance of instance-wise explainers on half moons data. Note that for the vast majority of points, LIME (orange) and EMAP (blue) provide near identical explanations (indicated by the direction of the arrows). Axis are arbitrary features $x_1$ and $x_2$. (Left): LIME $x_1$, $x_2$ coefficients plotted as vectors starting at the location of the point to be explained. (Right): EMAP negative perturbations plotted as vectors starting at the location of the point to be explained. Smallest 10\% of EMAP vectors indicated in red.}
    \label{fig3}
\end{figure*}

where $\tau$ is a hyperparameter governing the temperature of the distribution; as it approaches 0, the Gumbel-Softmax distribution approaches the Categorical distribution. $g_k \thicksim -\log(-\log(U))$, where $U \thicksim \text{UNIFORM}(0,1)$. This path derivative estimator allows us to backpropagate through the parameters of the sample for each categorical variable and thus train the perturbation model $g$ (see \citet{jang2016categorical} for more details).

\par When training $g$, these samples are then concatenated into a perturbed instance, $\pmb p^{(n)}$, which is passed through the pre-trained surrogate model $s(\pmb p^{(n)}; \hat{\theta}_s)$ as before. 

\par The only other difference to the training procedure is that the term in the objective intended to minimise the size of the adversarial perturbations in (3), $\lambda|g(\pmb x^{(n)};\theta_g)|_2$, must be changed to account for the fact we are no longer perturbing by adding small vectors to an input in $\mathbb{R}^d$. We make the simplest assumption that if perturbed feature $p^{(n)}_i$ takes on the same value as the original feature $x^{(n)}_i$, it has a perturbation cost of 0, and otherwise has a cost proportional to a hyperparameter $\eta$. This yields the following regularisation term:

\begin{equation}
    \text{reg}(\pmb x^{(n)}) =  \eta \sum_i^{D} \frac{1}{2}\Big|x^{(n)}_i - p^{(n)}_i\Big|
\end{equation}

Where $x^{(n)}_i$ and $p^{(n)}_i$ are 1-hot vectors of length $K$ (which may be different for different $i$), and $D$ is the number of categorical variables in $\pmb x$.

\par Our approach also supports a hybrid of both categorical and continuous variables, by combining the two objectives outlined above, where each affects the appropriate variables. The main challenge here is the relative magnitudes of $\lambda$ and $\eta$. We found (see discussion in Results, below), that for simple datasets setting $\lambda$ to around an order of magnitude smaller than $\eta$ yielded good results.

\section{Experimental Setup}

For all experiments below, expect otherwise stated, the neural network parameterising $g(\pmb x; \theta_g)$ consists of four fully connected layers of size 100 with ReLU nonlinearities and a 'partial gumbel layer' that combines standard additive perturbations for continuous variables with a collection of Gumbel-Softmax outputs for categorical variables, as discussed in the 'Model' section, above. We used a dropout percentage of 20 for every layer.

The neural network parameterising the surrogate $s(\pmb x; \theta_s)$ consists of three fully connected layers of size 200, with the first two nonlinearities being ReLU, and the final Softmax. We used a cross-entropy loss, as is standard for classification, and trained both models using Adam \citep{kingma2014adam}, with a learning rate of 1e-3. On simple synthetic datsets, both networks converge in under 15 epochs.

Network architecture and hyperparameters were chosen to be as simple as possible whilst providing reasonable results on a variety of datasets. Our intention was to showcase the generality and robustness of our model, so we avoided hyperparameter tuning or intensive model selection. Several similar architectures (more layers, wider fully connected layers) worked equally well, and an analysis of their relative merits is not pertinent to this initial presentation of the model.

For simple synthetic data we used 10000 samples from the half moons dataset, available on scikit-learn \citep{scikit-learn}, with Gaussian noise with standard deviation 0.2 added to the data. Our second synthetic dataset was handcrafted, and is described in the Results section, below. A more realistic continuous dataset was MNIST \citep{lecun1998gradient}, which we converted to a binary classification task by using only the digits 8 and 3, which gave a train/test split of 11982/1984, and training a classifier to predict between them. This approach was followed by both \citet{lundberg2017unified} and \citet{chen2018learning}.

Finally, to test the performance of our method on a mix of categorical data and continuous data, we used a dataset available from the UCI machine learning repository \citep{Dua:2019}. This was a subset of the Adult dataset, where in a similar fashion to \citet{white2019measurable} uninformative or highly skewed features ('fnlwgt', 'education', 'relationship', 'native-country', 'capital-gain', 'capital-loss') were removed, along with instances with missing values. The two classes were then balanced by undersampling the larger class, yielding a 17133/5711 train/test split. This left 3 continuous features, which were normalised to have zero mean and unit variance, and 5 categorical features (see Table \ref{table1} for example instances).

\section{Results}

\subsection{Continuous Features - Comparison to LIME}
We first demonstrate that in simple continuous input spaces, EMAP closely approximates LIME on standard a synthetic dataset, and succeeds in highlighting regions of interest in a manner unavailable to LIME.

We trained a Random Forest with 200 trees to classify the half-moons dataset (with a train/test split of 8000/2000) provided as standard with the scikit-learn toolset \citep{scikit-learn}. The classifier had an f-score of 0.97 on the test set. We then generated explanation coefficients for the classification 2000 randomly sampled points in the dataset using the off-the-shelf LIME toolkit \citep{ribeiro2016should}. Figure \ref{fig3}(left) shows 750 of these coefficients plotted as vectors starting at the location of the point to be explained.

\begin{figure}[ht]
    \centering
    \includegraphics[width=0.8\columnwidth]{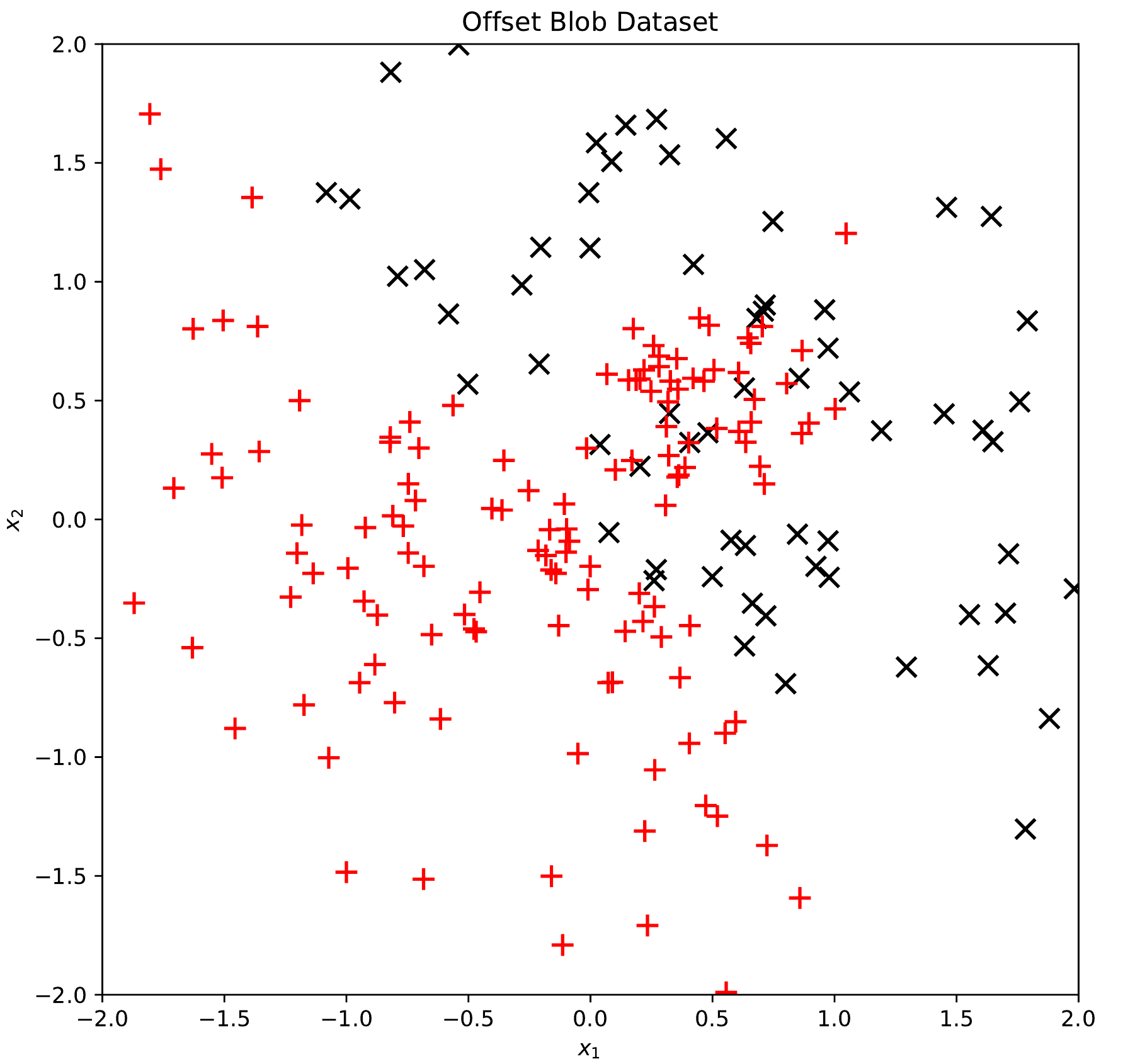}
    \caption{Example handcrafted 'offset blob' data. Class 1: red points, Class 2: black points. This dataset was used to showcase EMAP's ability to highlight regions of high classification uncertainty.}
    \label{fig:2}
\end{figure}{}

Secondly, we trained our surrogate on the input/output pairs of the Random Forest classifier, again with a 8000/2000 split. Our surrogate achieved a recovery accuracy of 0.981 on the test set\footnote{It recovered the Random Forest's classification 98\% of the time.}. We then trained our perturbation network on the opposite class labels, and it achieved an adversarial accuracy of 0.977 on the test set. Figure \ref{fig3}(right) shows the negative minimal perturbations returned by the perturbation network for the same 750 points explained by LIME. 

We present the negative perturbations for ease of comparison - by construction, minimal perturbations will point towards the nearest decision boundary whilst the weights of LIME's fitted logistic regression will point away from the nearest decision boundary. If presenting EMAP's outputs as explanations of the actual classification \textit{a la} LIME, this negation is necessary. If presenting EMAP's outputs as the perturbations required to cause a \textit{miss-classification}, the outputs of the perturbation network can be directly reported.

In this simple continuous space, the explanations output by EMAP correspond closely with those output by LIME. The mean cosine similarity between the 2000 LIME explanations and the 2000 EMAP explanations is 0.936. 

In addition, EMAP has two clear advantages over LIME on this sort of data; it is faster, and it indicates how close an instance is to a decision boundary, which can be treated as a proxy to how confident we should be in the black box classifier's prediction. In terms of speed, the time for LIME to generate the 2000 explanations above was 214 seconds. EMAP took 53.5 seconds to train once, and subsequently generated 2000 explanations in 1.32e-2 seconds.

\begin{figure}[ht]
    \centering
    \includegraphics[width=0.8\columnwidth]{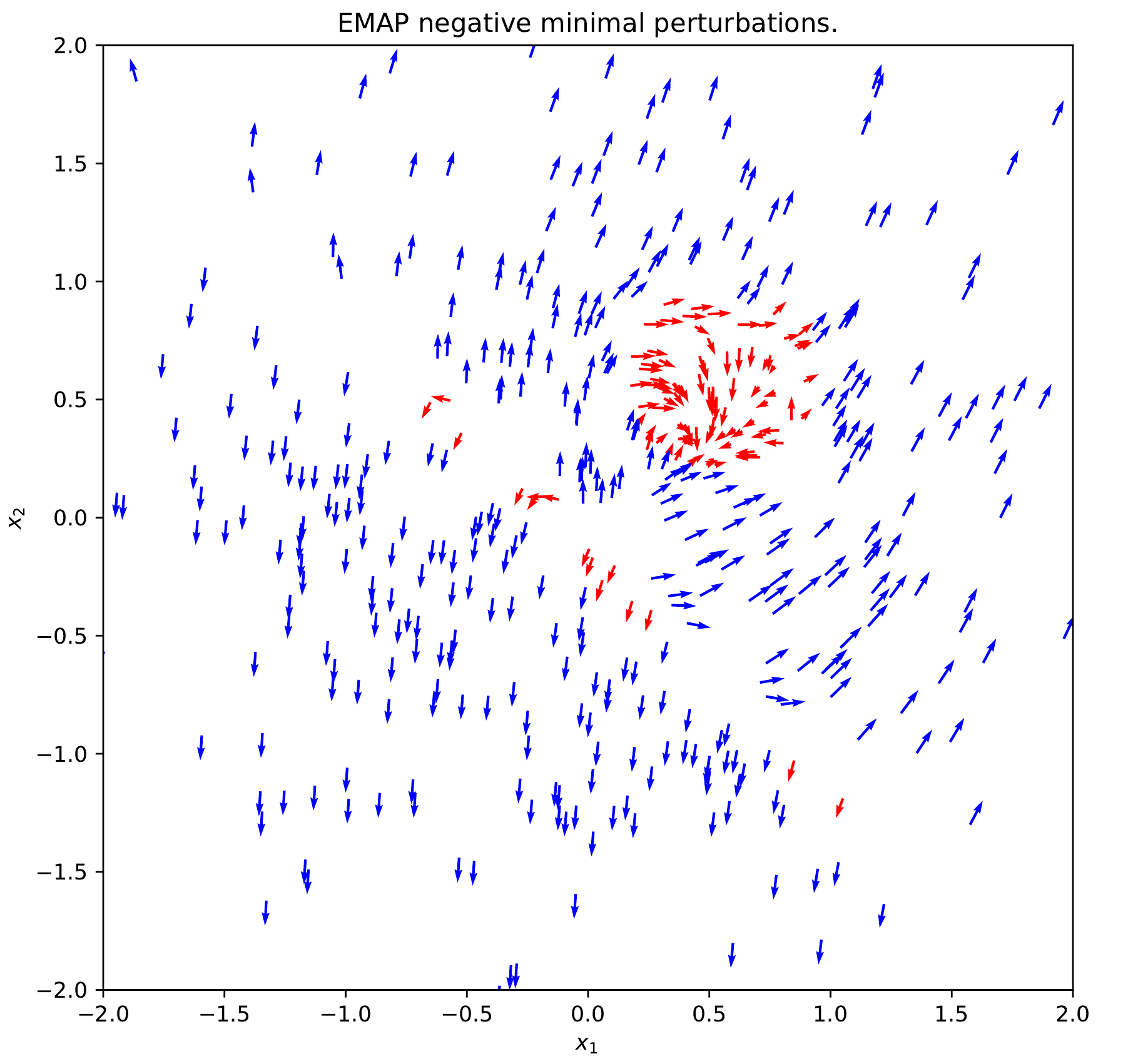}
    \caption{Example 'Offset blob' explanation directions from EMAP. Smallest 20\% of negative perturbation vectors in red, indicating regions of high classification uncertainty. Arrow lengths have been normalised.}
    \label{fig:3}
\end{figure}{}

As a consequence of regularising to return the \textit{minimal} perturbation instances which have perturbations with small magnitude, relative to the average for the dataset, are instances close to a decision boundary. This might be an indication that these instances are worth further examination; either by a preferred but slower explanation model, or directly by a user attempting to diagnose the behaviour of the black box model. In Figure \ref{fig3}(right), the smallest 10\% of perturbation vectors have been highlighted in red, and clearly track the decision boundary. Removing them from the cosine comparison improves mean cosine similarity with LIME's explanations to 0.964.

\begin{figure*}[h]
    \centering
    \includegraphics[width=\textwidth]{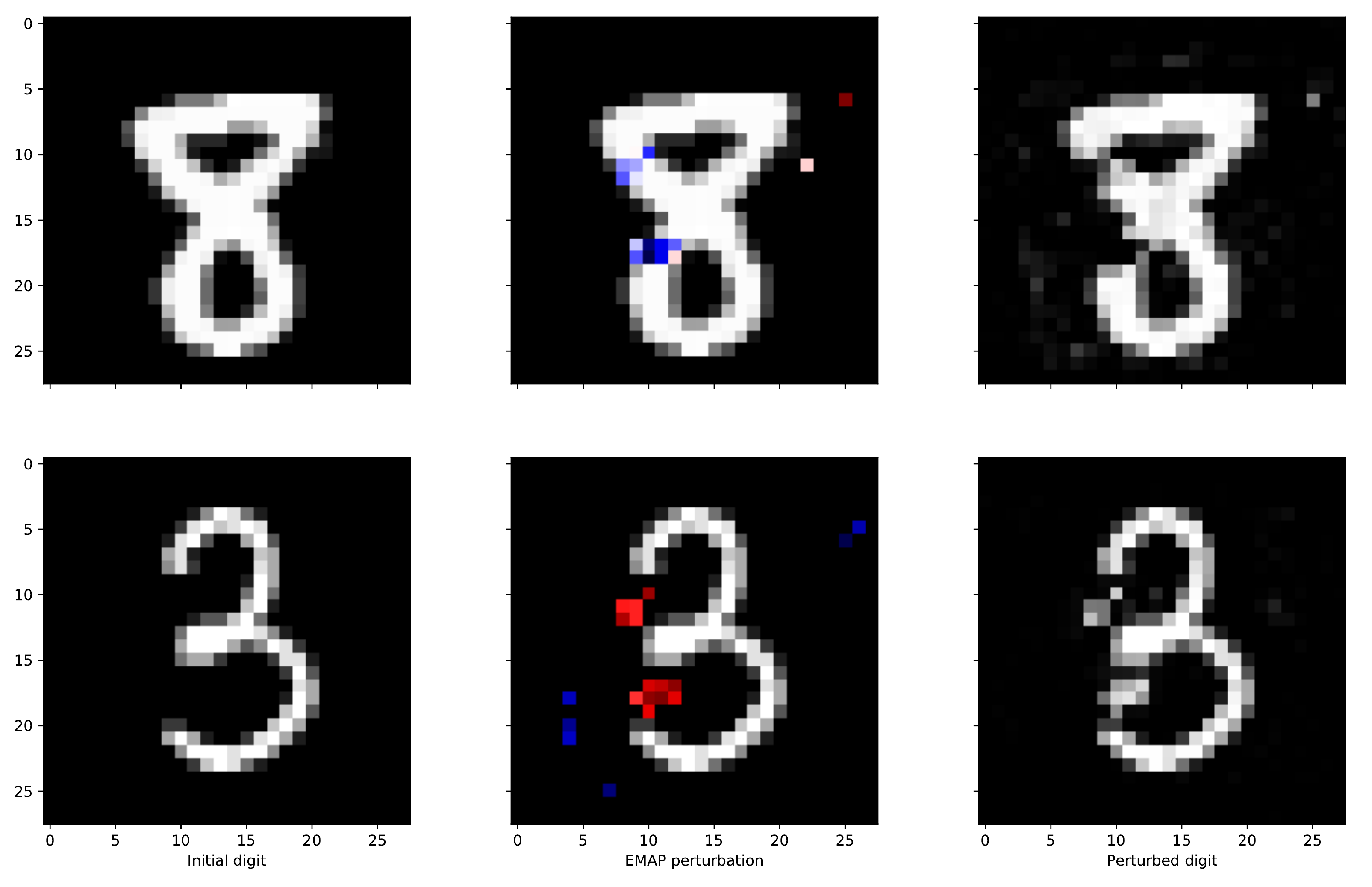}
    \caption{EMAP perturbations to MNIST digits at the pixel level. (Top): minimal perturbation to flip the Random Forest's classification from '8' to '3', and surrogate softmax output from [0.9902, 0.0098] to [0.000, 1.000]. (Bottom): minimal perturbation to flip Random Forest's classification from '3' to '8', and surrogate softmax output from [0.000,1.000] to [0.8910, 0.1090]. For the central images, blue indicates pixels which are substantially reduced in value, red pixels substantially increased in value. (Substantially means an increase or decrease of $>$ 0.2, where MNIST pixel values have been scaled to lie in [0,1]).}
    \label{mnistfig}
\end{figure*}{}

That this functionality has the potential to highlight regions of interest can be demonstrated using a simple handcrafted dataset, which we have called the 'offset blob' dataset (see Figure \ref{fig:2}). Here, a binary classification problem with a simple linear boundary is complicated by a region of positive instances within the general negative region. Data is generated from a standard 2d Normal distribution, and classified as class 2 if $x_1 \geq -x_2$, and class 1 otherwise. Additionally, a smaller amount of data (20\%), all class 1, is generated by $\thicksim N(\mu_\epsilon, \sigma_\epsilon)$, where $\mu_\epsilon = [0.5,0.5]$ and $\sigma_\epsilon = [0.25,0.25]$. The intention was to simulate a dataset where the black box classifier's decision boundary will necessarily be somewhat uncertain in a particular region.

As before, we trained a Random Forest classifier on the synthetic dataset, and it achieved an f-score of 0.910. We then trained our surrogate and perturbation network on the Random Forest's classifications, with a recovery accuracy of 0.935 on the test set, and an adversarial accuracy of 0.921 on the test set, respectively. The lower recovery and adversarial accuracies might be an indication that something is amiss - mean cosine similarity between 1100 LIME explanations and 1100 EMAP negative perturbations is also substantially lower, at 0.688.

Figure \ref{fig:3} shows the explanation vectors generated by EMAP for the critical region of the space. Note that whilst the minimal perturbations given for points in the unclear region are partly incorrect (as the region contains instances from both class 1 and class 2, there is no single 'true' solution for EMAP or the underlying black box model to find) they are also extremely small - to the extent that we were forced to normalise the vector lengths in Figure \ref{fig:3} to make them visible. 

In comparison, LIME's explanations of the points in the critical region are uninformative. LIME's explanations of class 1 points in the critical region are identical in direction to those of the in the general class 1 region - LIME's linear fit to sampled data for those points captures the general trend of the data only; and fails to indicate that there is anything amiss. Thus we might characterise LIME's output for a point in the critical region as correct but misleading, whereas EMAP's output is incorrect but indicative of low confidence. 

\begin{table*}[h]
\begin{tabular}{lllllllll|l}
\hline
                                                                   & Age  & \begin{tabular}[c]{@{}l@{}}Education\\ Years\end{tabular} & \begin{tabular}[c]{@{}l@{}}Weekly\\ Hours\end{tabular} & \begin{tabular}[c]{@{}l@{}}Work Class\\ Type\end{tabular} & \begin{tabular}[c]{@{}l@{}}Marital\\ Status\end{tabular} & Occupation                                                      & Race  & Sex    & \begin{tabular}[c]{@{}l@{}}Model\\ Class\end{tabular} \\ \hline
\begin{tabular}[c]{@{}l@{}}Example 1:\\ True Features\end{tabular} & 41   & 10                                                        & 60                                                     & Private                                                   & Married                                                  & \begin{tabular}[c]{@{}l@{}}Executive/\\ Management\end{tabular} & White & Male   & \textgreater{}\$50K                                   \\
Perturbation                                                       & 34.8 & 9.61                                                      & 40.0                                                   & Private                                                   & Widowed                                                  & \begin{tabular}[c]{@{}l@{}}Farming/\\ Fishing\end{tabular}      & White & Male   & \textless{}=\$50K                                     \\ \hline
\begin{tabular}[c]{@{}l@{}}Example 2:\\ True Features\end{tabular} & 40   & 9                                                         & 40                                                     & \begin{tabular}[c]{@{}l@{}}Local\\ Gov\end{tabular}       & \begin{tabular}[c]{@{}l@{}}Never\\ Married\end{tabular}  & \begin{tabular}[c]{@{}l@{}}Other/\\ Service\end{tabular}        & White & Male   & \textless{}=\$50K                                     \\
Perturbation                                                       & 43.0 & 11.5                                                      & 45.9                                                   & Private                                                   & Married                                                  & \begin{tabular}[c]{@{}l@{}}Executive/\\ Management\end{tabular} & White & Male   & \textgreater{}\$50K                                   \\ \hline
\begin{tabular}[c]{@{}l@{}}Example 3:\\ True Features\end{tabular} & 33   & 9                                                         & 35                                                     & Private                                                   & \begin{tabular}[c]{@{}l@{}}Spouse\\ Absent\end{tabular}  & \begin{tabular}[c]{@{}l@{}}Other/\\ Service\end{tabular}        & Asian & Female & \textless{}=\$50K                                     \\
Perturbation                                                       & 46.5 & 14.1                                                      & 43.0                                                   & Private                                                   & Married                                                  & \begin{tabular}[c]{@{}l@{}}Admin/\\ Clerical\end{tabular}       & White & Male   & \textgreater{}\$50K                                   \\ \hline
\end{tabular}
\caption{Example instances (True Features) and their perturbations generated by EMAP on a subset of the UCI Adult dataset. Model Class indicates the classification assigned by the underlying Random Forest classifier.}
\label{table1}
\end{table*}

\subsection{Continuous Features - MNIST Pixel Perturbation}

To demonstrate EMAP's performance on more complex data with much larger feature spaces, we trained a 200 tree Random Forest classifier on a two-class subset of the MNIST dataset, where the classes were '8' and '3'. The Random Forest achieved an f-score of 0.98. We then trained EMAP on the label provided by the Random Forest. The structure of both surrogate and perturbation networks was identical to that in the simple synthetic cases detailed above (see Data and Methods section for an overview). The surrogate model achieved a recovery accuracy of 0.983 on the test set, and the perturbation network an adversarial accuracy of 0.975 on the test set.  

Figure \ref{mnistfig} shows examples of the minimal perturbation required to change the surrogate's classification to the incorrect label. Both instances shown also flip the classification of the unseen Random Forest. As can be seen, EMAP has learned to either remove part of the left hand strokes of 8s, or partly fill the gaps for 3s. That it does not do so fully is due to its remit to recover minimal perturbations - it does not need to fully remove or redraw the relevant part of the letter to flip the classifier's decision. 

\subsection{Categorical Features}

Lastly, we show how EMAP handles a mixture of continuous and categorical variables. On a subset of the UCI Adult Dataset \citep{Dua:2019}, our Random Forest achieved an f-score of 0.81, and EMAP a surrogate accuracy of 0.882, and an adversarial accuracy of 0.853.

Table \ref{table1} shows three example perturbations produced by EMAP. An open question when dealing with data with a mixture of variables is: to what extent are perturbations comparable? In Example 1, Table \ref{table1}, EMAP reduces the age of the individual by around 6 years, and changes their marital status from 'Married' to 'Widowed'. Which is a more substantial change? When searching for a \textit{minimal} perturbation, the model's relative weighting of age (a continuous variable) and marital status (a discrete variable) is dependant on the value ascribed to the relative magnitudes of $\lambda$ and $\eta$, the hyperparameters weighting the minimsing regularisation terms for continuous and discrete variables, respectively (see Equation (5)).

In practice, we found that it was necessary to set $\eta$ to around an order of magnitude larger than $\lambda$ (the values for the above perturbations were $\eta = 2.0$, $\lambda = 0.1$), to prevent the model from making such substantial changes to the categorical variables of each instance so as to be uninformative. With this setting, we found that changes to marital status and occupation dominated the minimal perturbations for those individuals who were already close to the boundary. Both Example 1 and Example 2 in Table \ref{table1}, for instance, remain white and male. However, Example 3 requires substantial changes to almost every variable to convince the classifier to change its decision.

More fine-grained analysis could involve comparing the values of the $\log \pi_k$ values passed to the gumbel softmax layer directly, and regularising these outputs. A second approach, in a setting where we had access to a learned embedding for categorical variables, might be to use the distance travelled in that embedding space to $p^{(n)}_i$ from $x^{(n)}_i$ as a proxy to size of perturbation. We intend to pursue these avenues in further work.

\subsection{Stability}
 One question about our method might be that because gradient-based optimization can lead to a local minimum, the outputs with respect to the same input, or two inputs with small changes can change drastically. 
 Our initial approach to using the differentiable surrogate was to use the gradients of the input to the surrogate directly; such that the direction of minimal perturbation required to flip the classifier was taken to be the local gradient of the input with respect to the negative loss. In development, however, we ran into exactly the problem described above - multiple minima lead to instability on retraining, and (particularly) along decision boundaries of the underlying classifier.
 
 We found that adding a second network trained to output the perturbations directly helped smooth these instabilities out substantially, particularly when the output of the second network was heavily regularised. (Having a second network is also slightly faster; we get an explanation with a single forward pass, rather than a forward pass and a backwards pass). For example, the mean cosine similarity between perturbations output by two runs of EMAP on the MNIST dataset is 0.9497.
 
 Combined with our approaches’ ability to highlight areas of potential instability, we consider it to demonstrate reasonable robustness, at least on the presented datasets. With regard to optimization algorithms, on our data we observe little difference as long as both networks train.

\section{Conclusion}

Whilst we have compared ourselves to the literature on model-agnostic instance-wise explanation, we are not necessarily in competition with it. EMAP can be thought of an additional tool in the model development toolbox; useful both for its speed, and its ability to indicate regions of data space where further investigation of the behaviour of the underlying classifier is warranted.

EMAP's speed is one of its primary assets; where sampling based explanation methods may be too slow to provide instance-wise explanations of a large dataset in a reasonable amount of time, once trained EMAP merely needs a single forward pass to output a perturbation vector. As data can be batched before input, EMAP can handle large numbers of instances rapidly. 

Secondly, EMAP can be thought of as a novel approach in that it proposes \textit{minimal adversarial perturbations} as a useful explanatory tool. This aligns it with the literature on counterfactuals as explanations \citep{wachter2017counterfactual}, as the minimal adversarial perturbation can also be thought of as the minimal \textit{counterfactual direction} - the direction in which one could perturb an instance to cause a classifier to change its classification. 

Thirdly, EMAP provides novel functionality with its ability to highlight regions of space of potential interest to a user, or that pose potential problems for the underlying classifier.

Finally, as we have shown in continuous feature spaces, EMAP also produces results comparable (under a change of sign) to the output of additive feature attribution methods, such as LIME and SHAP \footnote{We should be clear that EMAP is not itself an additive feature attribution method.}. This means that we can think of EMAP as an empirical demonstration of the relationship between two distinct paradigms of explanation; that the vector of feature contributions to the output of some model $f$ for some instance $\pmb x^{(n)}$ is the negative of the direction of perturbation to that instance required to recover its nearest counterfactual $\pmb p^{(n)}$.

\bibliography{refs}
\bibliographystyle{aaai}
\end{document}